\documentclass[letterpaper, 10 pt, conference]{ieeeconf}  

\IEEEoverridecommandlockouts                              

\usepackage{algorithmic}
\usepackage{graphicx}
\usepackage{textcomp}
\usepackage{xcolor}
\usepackage{color, colortbl}
\definecolor{Gray}{gray}{0.9}
\definecolor{LightCyan}{rgb}{0.88,1,1}
\newcommand{\mycomment}[1]{}

\title{\LARGE \bf
Object Detection Approaches to Identifying Hand Images with High Forensic Values
}

\author{Thanh Thi Nguyen$^{1}$, Campbell~Wilson$^{1}$, Imad~Khan$^{1}$ and Janis~Dalins$^{2}$
\thanks{$^{1}$Thanh Thi Nguyen, Campbell Wilson and Imad Khan are with the AiLECS Lab, Faculty of Information Technology, Monash University, Melbourne, VIC 3800, Australia
        {\tt\small \{thanh.nguyen9,campbell.wilson\}@monash.edu}}%
\thanks{$^{2}$Janis Dalins is with the AiLECS Lab, Australian Federal Police,
        Melbourne, VIC 3800, Australia
        {\tt\small janis.dalins@afp.gov.au}}%
}

\begin{document}

\maketitle
\thispagestyle{empty}
\pagestyle{empty}

\begin{abstract}

Forensic science plays a crucial role in legal investigations, and the use of advanced technologies, such as object detection based on machine learning methods, can enhance the efficiency and accuracy of forensic analysis. Human hands are unique and can leave distinct patterns, marks, or prints that can be utilized for forensic examinations. This paper compares various machine learning approaches to hand detection and presents the application results of employing the best-performing model to identify images of significant importance in forensic contexts. We fine-tune YOLOv8 and vision transformer-based object detection models on four hand image datasets, including the 11k hands dataset with our own bounding boxes annotated by a semi-automatic approach. Two YOLOv8 variants, i.e., YOLOv8 nano (YOLOv8n) and YOLOv8 extra-large (YOLOv8x), and two vision transformer variants, i.e., DEtection TRansformer (DETR) and Detection Transformers with Assignment (DETA), are employed for the experiments. Experimental results demonstrate that the YOLOv8 models outperform DETR and DETA on all datasets. The experiments also show that YOLOv8 approaches result in superior performance compared with existing hand detection methods, which were based on YOLOv3 and YOLOv4 models. Applications of our fine-tuned YOLOv8 models for identifying hand images (or frames in a video) with high forensic values produce excellent results, significantly reducing the time required by forensic experts. This implies that our approaches can be implemented effectively for real-world applications in forensics or related fields.

\end{abstract}


\section{INTRODUCTION}
\label{sec_int}

The integration of machine learning-based object detection methods in identifying hand images with high forensic values holds significant promise in enhancing the efficiency and accuracy of forensic investigations. Forensic studies suggest that these images may have relevance in legal or investigative matters \cite{hackman2023forensic}. This could include identifying individuals based on hand features or using hand images as evidence in forensic analysis. Measuring the forensic value of hands involves assessing various features and characteristics that can aid in identification or analysis within forensic contexts. Aspects that contribute to the forensic value of hands may include fingerprints, palmprints, hand geometry, vein patterns, scars and tattoos, or occupational markers. 

\emph{Fingerprints} are distinctive ridge patterns on fingers that can be used for authentication and verification needs, making them one of the most valuable forensic features of hands. Fingerprints are unique to each individual and have been a traditional method for personal identification \cite{singla2020automated}. Automated systems for fingerprint recognition can capture and analyze the ridge patterns and minutiae points present on an individual's fingertips. Similar to fingerprints, \emph{palmprints} also have unique ridges or distinctive patterns. Palmprint recognition systems can analyze the palm's surface for identification purposes \cite{badiye2023palmprints}. Likewise, \emph{hand geometry} involves measuring and analyzing physical characteristics of the hand, such as length and width of fingers, distance between joints, palm width, and overall shape of the hand \cite{Zhang2021}. Measurements of hand dimensions can provide valuable forensic information, particularly in cases where other biometric features are unavailable.

\emph{Vein patterns} in the hand, typically those visible on the back of the hand, can also be used for individual recognition, especially in conjunction with other biometric identifiers. The veins in the hand form a unique pattern that can be captured using infrared technology. Vein pattern recognition is a non-intrusive method and can be implemented as an effective identification tool \cite{kuzu2022intra}. Similarly, \emph{birthmarks} and \emph{tattoos} on the hands can serve as identifying markers and may offer significant forensic insights across various cases \cite{dahal2023interdisciplinary}. Certain occupations or activities may leave distinctive marks on the hands (i.e., \emph{occupational markers}), such as corns, calluses or scars, which are useful for identification or event reconstruction \cite{soltani2019formal}.

Artificial intelligence technologies such as object detection methods \cite{zou2023object} can be used to locate and isolate fingerprints, palm prints, or other unique features within a hand image. The technologies can assist forensic experts in quickly processing large volumes of image and video data, reducing the time required for manual analysis. 
Every day, law enforcement officers encounter digital video evidence, often necessitating the examination of substantial data volumes. This task can expose them to distressing content, including materials related to child sexual abuse and exploitation. Our overarching objective is to extract forensic value of hand images using artificial intelligence, mirroring the processes conducted manually by humans. The initial stride toward achieving our objective involves building a real-world, noise-tolerant automated hand recognition system. 

This paper presents a comparison between object detection approaches to identifying hand images with high forensic values, aiming to support forensic experts in their daily practice. Many modern object detection methods use learning-based approaches, such as deep learning models. Convolutional Neural Networks (CNNs) \cite{dhillon2020convolutional} and more advanced models like faster R-CNN \cite{sahin2023detection}, You Only Look Once (YOLO) \cite{diwan2023object}, single shot multibox detector \cite{wang2023single}, and Vision Transformers (ViT) \cite{gehrig2023recurrent} have demonstrated effectiveness in detecting various objects in an image. These models are trained on large datasets to learn the complex patterns and features associated with the objects of interest. 

In this study, we implement and compare performance of popular YOLO and modern ViT models for hand detection. More specifically, we employ two YOLOv8 variants, i.e., YOLOv8 nano (YOLOv8n) and YOLOv8 extra large (YOLOv8x), and two ViT variants, i.e., DEtection TRansformer (DETR) \cite{carion2020end} and Detection Transformers with Assignment (DETA) \cite{ouyang2022nms} for experiments using four hand image datasets. We then apply the best-performing approach to identifying images (or frames in a video) that possess high forensic values.
We maintain a high tolerance for false positive rates as these images and frames are subject to manual scrutiny by forensic experts afterward. Images or frames containing a substantial hand portion are deemed to have a high forensic value. The hand portion's size or area within an image can be calculated straightforwardly using the dimensions of the bounding boxes detected by the object detection models.

In summary, this work's contribution is threefold: 1) create a combined dataset of a wide range of hand images, which is helpful for effectively training hand detection models. In particular, we create bounding boxes on our own and make them available for every image in the 11k hands dataset \cite{afifi201911k}; 2) evaluate the performance of various object detection models on hand image datasets and identify the best-performing method; 3) implement the best-performing approach for identifying hand images with high forensic values, thereby significantly reducing the time required by forensic experts.

It is crucial to highlight that this research does not propose using any elements to identify or store individuals' sensitive or private information, including biometric data. It does not involve collecting new data or data beyond what is already legally accessible to law enforcement. The purpose of this work is solely to rank images based on their potential value to forensic examiners, adhering to established legal processes and controls.

\section{RELATED WORK}
\label{sec_pro_app}
In forensic contexts, offenders often exhibit forensic awareness by deliberately excluding their faces from images. However, there is less concern about whether other parts of their anatomy are visible, possibly due to the belief that identification is less likely from these areas. These body parts may encompass feet, legs, thighs, genitals, and abdomen. Notably, it is most often the hand (specifically the back of the hand) and forearm that are captured \cite{hackman2023forensic}. Hand images thus serve as a valuable resource for forensic analyses.

Hand analysis has been a subject of enduring interest in the field, with considerable research dedicated to aspects such as grasp analysis \cite{chao2021dexycb}, pose estimation \cite{spurr2021self} and reconstruction~\cite{lin2021end}. Nevertheless, these approaches have predominantly concentrated on controlled in-lab settings, frequently assuming a pre-localized hand or operating in environments with restricted variability. Despite substantial advancements, deploying these methods in the expansive realm of Internet videos presents a challenge, mainly attributed to the overwhelming diversity in viewpoints and contexts. The objective of the research presented in \cite{shan2020understanding} is to facilitate hand analysis on a large and diverse scale across the Internet. In pursuit of that goal, they propose a model capable of identifying various attributes for each individual hand in a given RGB image, demonstrated across a wide range of scales and contexts. These attributes include a bounding box for the hand, its orientation (left/right), contact state, and more. These identified attributes play a pivotal role in addressing downstream challenges such as pose reconstruction and grasp analysis.

Joshi and Kanphade present a forensic approach in \cite{joshi2020deep} using a multiple convolutional layer network along with a fully connected and a k-NN layer to identify a person. However, their approach requires a biometric radiograph, which is not readily available in most real-world law enforcement scenarios. In another work, Narasimhaswamy et al. \cite{narasimhaswamy2019contextual} introduced Hand-CNN, a model based on the CNN architecture for detecting hand masks and projecting hand orientations in unconstrained images. Hand-CNN enhances MaskRCNN \cite{he2017mask} by using a novel attention mechanism to integrate contextual cues in the detection procedure. They also created a large-scale hand dataset consisting of hands in unconstrained images with annotations, which are helpful for training and evaluating various machine learning models in this domain. 

On the other hand, a resilient hand tracking approach that combines a correlation filter with a correction strategy utilizing a fast object detection model, specifically the single-shot detection algorithm, is introduced in \cite{haji2023vision}. This amalgamation enables the tracker to reinitialize when hand movement is inaccurately traced, ensuring consistent and precise tracking. The approach minimizes computational costs of the detector by detecting the object of interest only during the initial frame and when it is mislocated by the tracker. This reduction in computational load leads to an enhancement in real-time performance.

\section{HAND DETECTION METHODS}
\subsection{YOLOv8 Approaches}
YOLOv8 \cite{yolov8_ultralytics} represents the newest generation within the YOLO-based object detection models by Ultralytics, showcasing cutting-edge performance. Building upon the advancements of previous YOLO versions, the YOLOv8 model offers enhanced speed and accuracy, presenting a unified framework for training models across: object detection, instance segmentation, and image classification.

YOLOv8 is the 2023 iteration in the YOLO series of models that features an architecture similar to YOLOv5, comprising a sequence of convolutional layers. Notably, YOLOv8 distinguishes itself by incorporating a cross-stage partial bottleneck in the convolutional layer. The output from these convolutional layers is subsequently directed to a decoupled head. This decoupling enables the head to independently focus on distinct tasks, encompassing object detection, classification, and regression. It is important to clarify that our work specifically pertains to object detection in this context.

We use YOLOv8 as one of the competing methods. There are several variants of YOLOv8, including nano, small, medium, large and extra large variants. We fine-tune the smallest variant, i.e., YOLOv8n, and the largest variant, i.e., YOLOv8x, for hand detection in this study. Among the YOLOv8 variants, YOLOv8n stands out as the quickest and most compact, whereas YOLOv8x distinguishes itself as the most accurate albeit the slowest.

\subsection{Vision Transformer Approaches}
We fine-tune Detection Transformer (DETR) \cite{carion2020end} and Detection Transformers with Assignment (DETA) \cite{ouyang2022nms} models for the hand detection applications. The variant of DETR used in this study is the facebook/detr-resnet-50, whereas that of DETA is the jozhang97/deta-swin-large. The pre-trained weights of these models are available on the Hugging Face repositories, respectively at https://huggingface.co/facebook/detr-resnet-50 and https://huggingface.co/jozhang97/deta-swin-large.

\subsubsection{DETR-ResNet-50}
This is a DETR model with a ResNet-50 backbone, which was pre-trained on the COCO 2017 object detection dataset including 118k images with annotations \cite{lin2015microsoft}. The DETR model employs an encoder-decoder transformer architecture with a convolutional ResNet model. To enable object detection, two heads are incorporated atop the decoder outputs: a linear layer responsible for predicting class labels and a multi-layer perceptron for bounding boxes. Employing object queries, the model searches for specific objects within an image, with each object query dedicated to locating a particular object.

The model undergoes a training process employing a ``bipartite matching loss'', wherein the inferred classes and bounding boxes of each object query are compared with annotated labels. These annotations are padded to match the length of the object queries, accommodating scenarios where an image comprises fewer objects. The Hungarian matching technique is then deployed to establish a best possible one-to-one mapping between each query and each annotated label \cite{detr-resnet-50}. Subsequently, in order to optimize the model parameters, the standard cross-entropy loss is employed for the classes, while a linear combination between L1 and generalized Intersection over Union (IoU) loss is applied for the bounding boxes. In our implementation, this model consists of 41.5M parameters in which 41.3M parameters are trainable and 222k parameters are non-trainable.

\subsubsection{DETA-Swin-Large}
The DETA model was proposed in \cite{ouyang2022nms}, which aims to rectify a commonly held misconception that one-to-one mapping is indispensable for achieving high-performance detection. Contrary to this belief, the study in \cite{ouyang2022nms} demonstrates that the conventional one-to-many training objective can produce equally adept detection transformers. The DETA approach involves crafting a transformer-based object detector that assigns positive-negative labels directly to each query, akin to traditional detectors. Additionally, they employ non-maximum suppression (NMS) method to eliminate redundant predictions, deviating from the conventional end-to-end one-to-one matching paradigm. The enhancement in this model involves substituting the one-to-one bipartite Hungarian matching loss, as employed in Deformable DETR \cite{zhu2021deformable}, with one-to-many label assignments akin to those used in conventional detectors, accompanied by the NMS mechanism. 
This model includes 218M parameters in total with all of them being trainable.

\section{DATASETS AND PERFORMANCE METRICS}
\label{sec_exp_dis}
\subsection{Used Datasets}
We employed four datasets to evaluate performance of the competing hand object detection methods. The first dataset is extracted from the EgoHands dataset \cite{bambach2015lending}, consisting of 4,800 images with 15,053 ground-truth labeled hands. These images are obtained from 48 videos, with 100 frames for each video. Preprocessing the data allows us to acquire 4,787 images that have available labels. Within these images, 3,590 images are selected randomly as the training data, while the validation and test sets include 335 and 862 images, respectively.

The second dataset is the 11k hands dataset \cite{afifi201911k}, comprising 11,076 hand images (1600$\times$1200 pixels) of 190 subjects, of varying ages between 18 and 75 years old. The third dataset is extracted from the Open Images dataset \cite{kuznetsova2020open, OpenImages2}, including 20,500 hand images for training, 1,892 images for validation, and 4,932 images for testing. 

To obtain a comprehensive comparison between the object detection approaches, we create a combined dataset that includes all aforementioned datasets. More specifically, the training and validation sets of the combined dataset include respectively 32,397 and 3,002 images, while the test set consists of 7,788 images. A statistical summary of these four datasets is presented in Table \ref{tab_data_stats}. 

\begin{table}[htbp]
\caption{Summary Statistics of Datasets}
\begin{center}
\begin{tabular}{|l|r|r|r|l|}
\hline
\textbf{Datasets} & \textbf{Train} & \textbf{Val} & \textbf{Test} & \textbf{Features} \\
\hline
\rowcolor{LightCyan}
EgoHands & 3,590 & 335 & 862 & Max 4 hands per image\\
\hline
11k hands & 8,307 & 775 & 1,994 & One hand per image \\
\hline
\rowcolor{LightCyan}
Open Images & 20,500 & 1,892 & 4,932 & Images in the Wild\\
\hline
Combined & 32,397 & 3,002 & 7,788 & All of the above combined\\
\hline
\end{tabular}
\label{tab_data_stats}
\end{center}
\end{table}

\begin{figure*}[ht]
    \frame{\includegraphics[width=.33\textwidth]{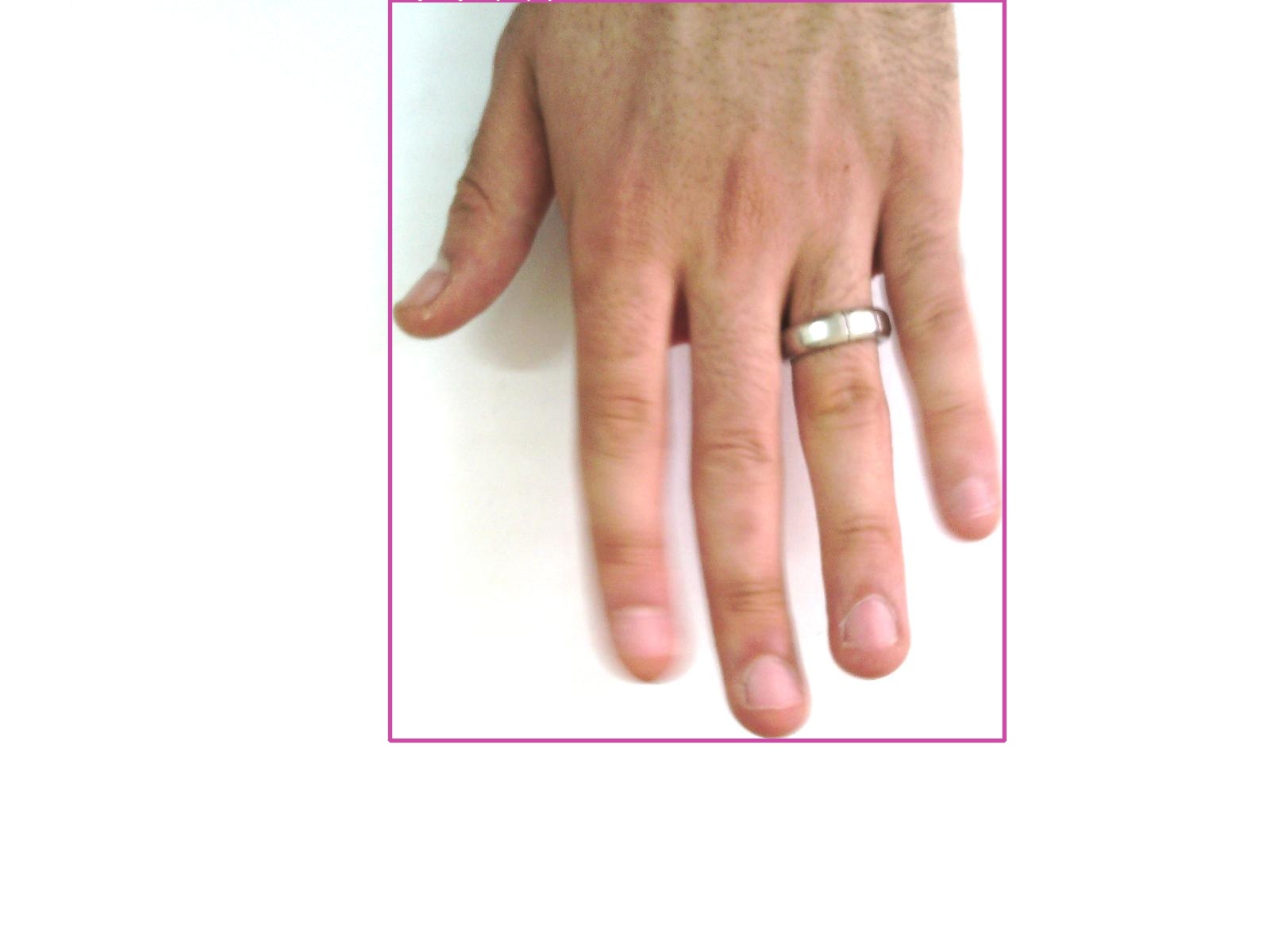}}\hfill
    \frame{\includegraphics[width=.33\textwidth]{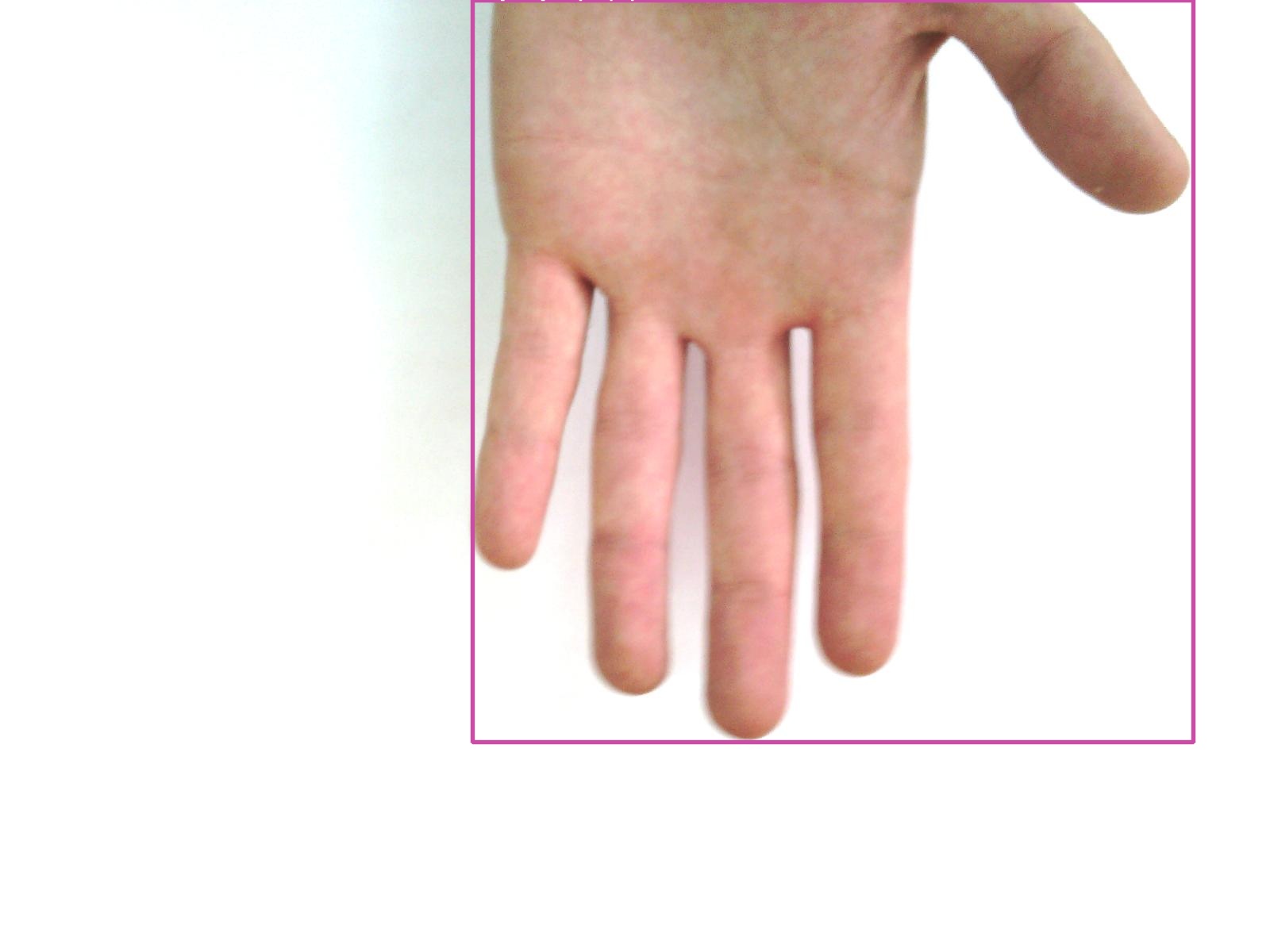}}\hfill
    \frame{\includegraphics[width=.33\textwidth]{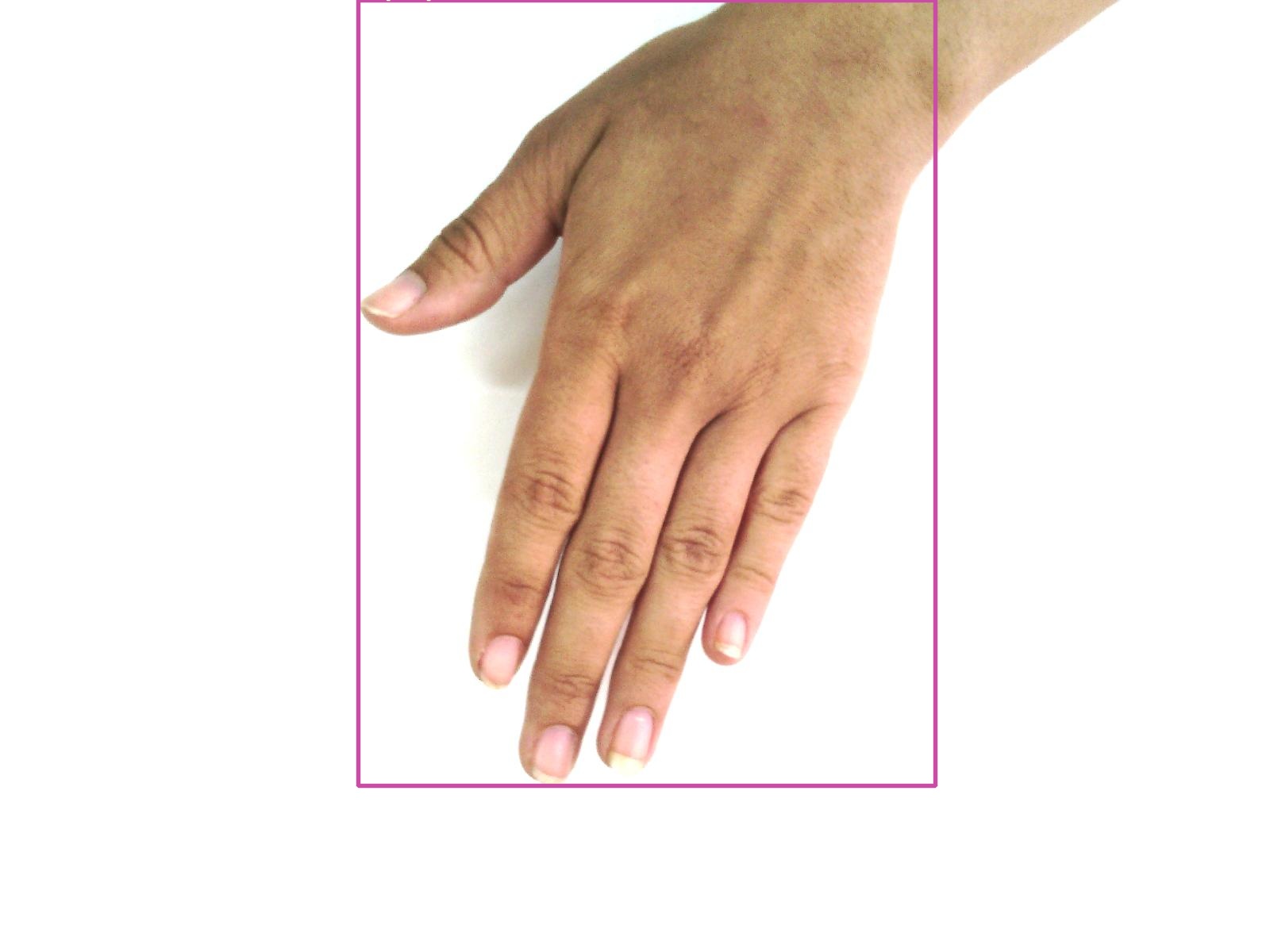}}
    \\[\smallskipamount]
    \frame{\includegraphics[width=.33\textwidth]{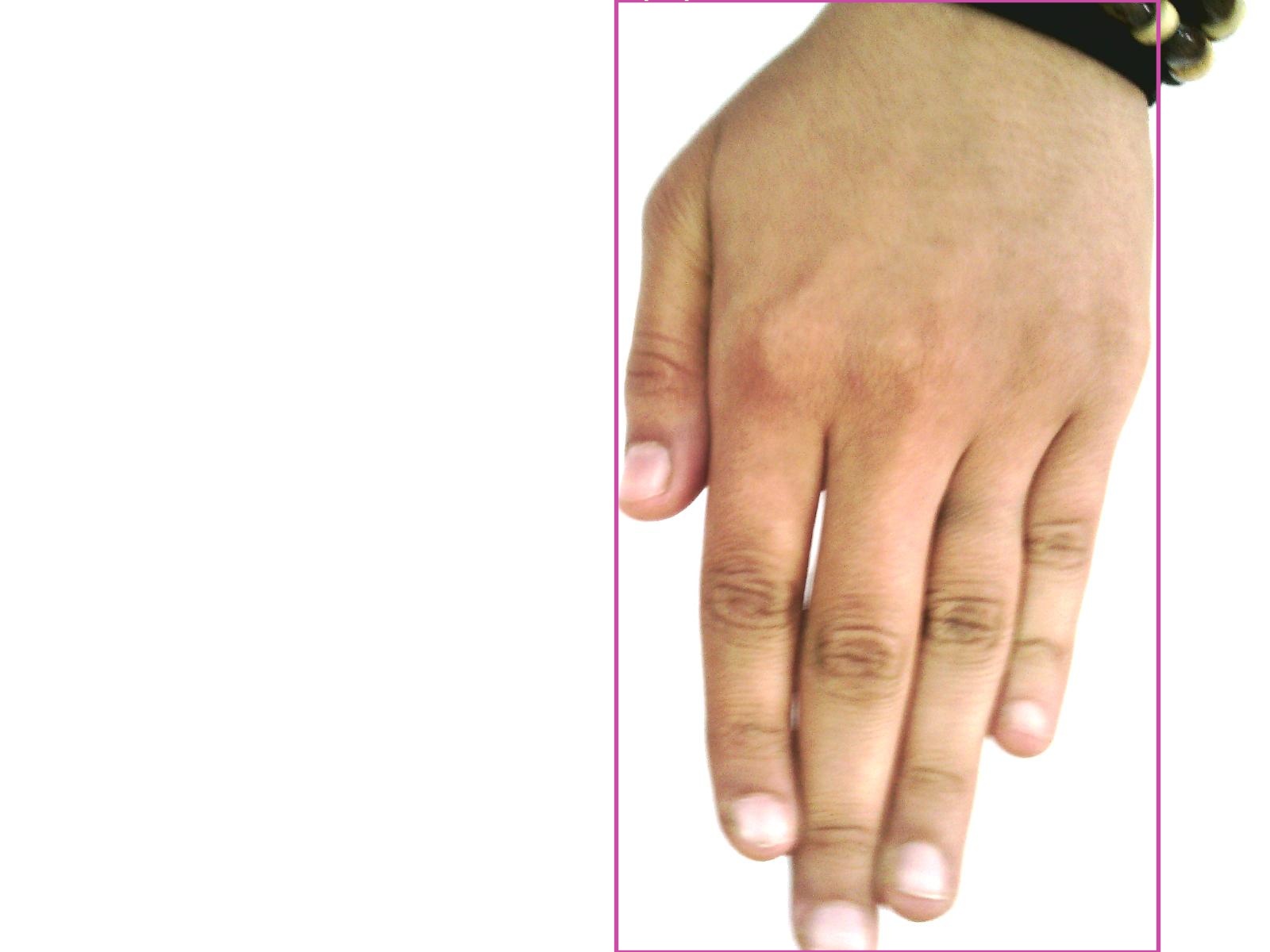}}\hfill
    \frame{\includegraphics[width=.33\textwidth]{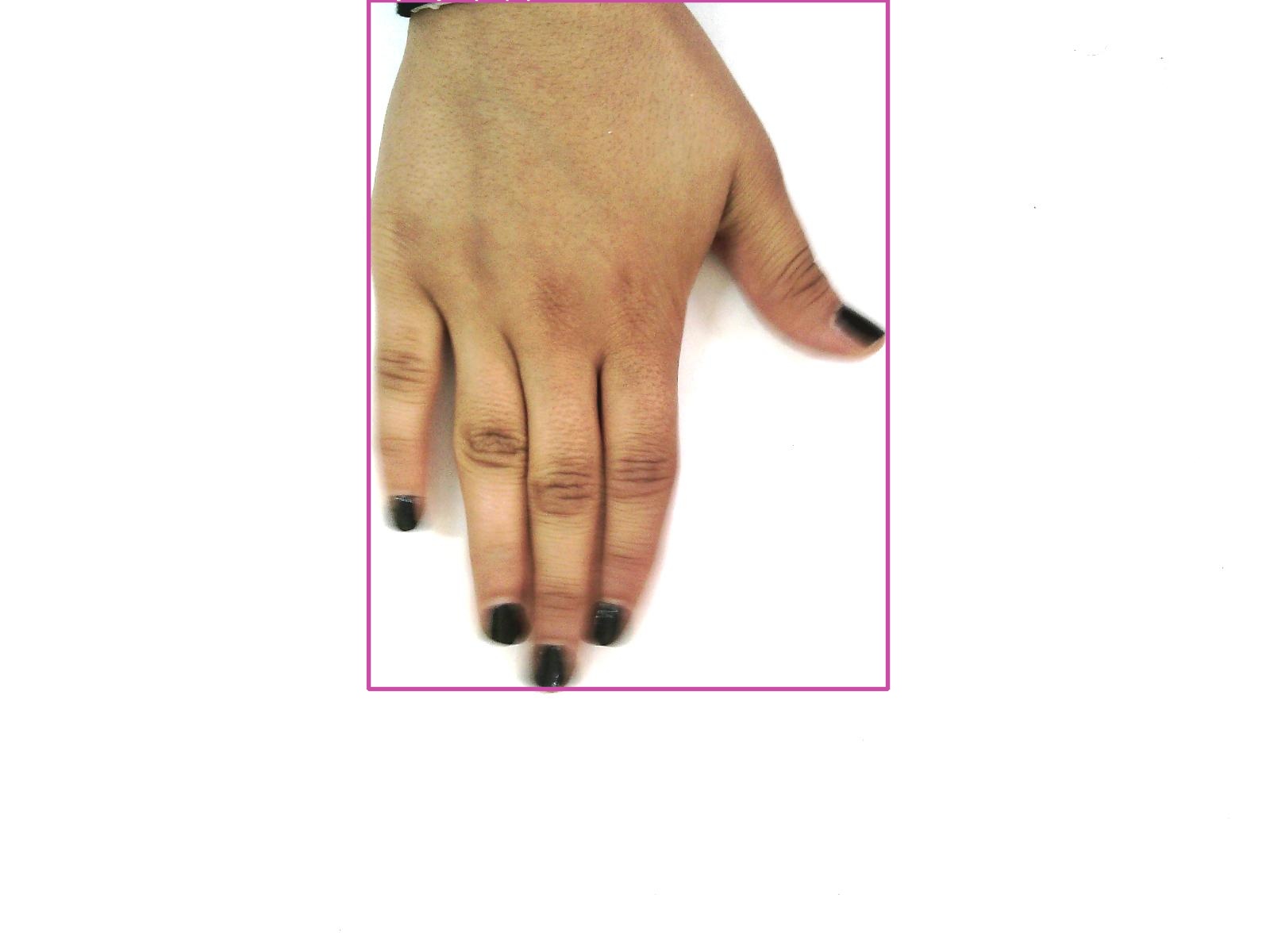}}\hfill
    \frame{\includegraphics[width=.33\textwidth]{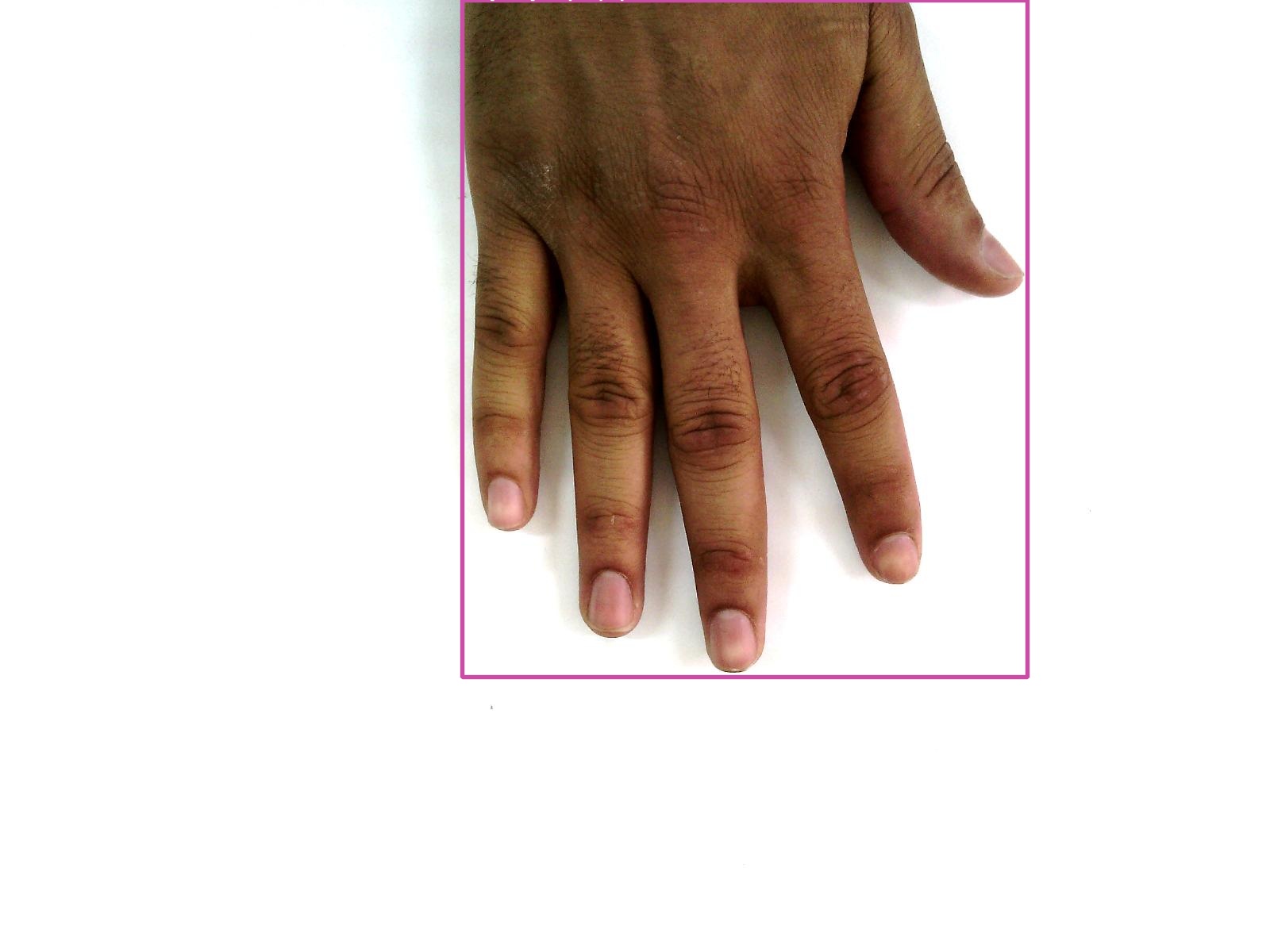}}
    \caption{Six example images from the 11k hands dataset \cite{afifi201911k} with our labels, which are bounding boxes in the pink color. These bounding boxes are in the YOLO format and made publicly available for every image in the 11k hands dataset.}
    \label{fig:11k_hands}
\end{figure*}

While the EgoHands and Open Images datasets have ground-truth bounding boxes available and can be obtained from their respective sources, the 11k hands dataset does not have bounding box labels and thus requires more involved processing steps to annotate its 11,076 hand images. As a contribution of this work, we have created and made bounding boxes publicly available and accessible for every image in the 11k hands dataset. Our approach is \emph{semi-automatic} as it combines both manual human efforts and automated procedures. We start by randomly selecting 500 images from this dataset and manually labeling these 500 images. We then implement the following four steps:

1) Train a rudimentary model by fine-tuning the YOLOv8n model on the 500-image dataset. 

2) Use the fine-tuned model to predict bounding box labels for the rest of the images from the 11k hands dataset.

3) Select 1,000 images with top confidence scores from the prediction results in Step 2). 

4) Add these 1,000 images to the 500-image dataset to make a new training dataset of 1,500 labelled images. 

We repeat the four steps above, but this time we fine-tune the YOLOv8n model on the 1,500-image dataset. This repeat results in a new training dataset of 2,500 labelled images. We then fine-tune the YOLOv8n model on this 2,500-image dataset and use the fine-tuned model to finally predict labels (bounding boxes) for the rest of the images from the 11k hands dataset. These predicted bounding boxes (i.e., for 8,576 images), along with those of the 2,500-image dataset, are considered as ground-truth labels for images in the 11k hands dataset used in this paper. Several images of this dataset are displayed along with their bounding boxes in Fig. \ref{fig:11k_hands}. All bounding boxes are in the YOLO format and made available at: https://github.com/thanhthinguyen/boxes-11k-hands.

\subsection{Performance Metrics}

Average Precision (AP) and Average Recall (AR) are popular metrics for evaluating the accuracy of object detection methods by estimating the Precision-Recall relationship \cite{open_od_leaderboard}. There are a number of variations of these metrics, depending on the Intersection over Union (IoU) threshold, the area of the object, and the number of objects per image. The most popular metrics are listed in Table \ref{tab_metrics}.

\begin{table}[htbp]
\caption{Summary of Object Detection Performance Metrics \cite{open_od_leaderboard, padilla2020survey}}
\begin{center}
\begin{tabular}{ll}
\hline
\textbf{Metrics} & \textbf{Descriptions} \\
\hline
\multicolumn{2}{l}{\textbf{Average Precision (AP)}} \\
\hline
\rowcolor{LightCyan}
AP & AP at IoU=.50:.05:.95 \\
\hline
AP@.50 (APIoU=.50) & AP at IoU=.50 ($\sim$ PASCAL VOC mAP) \\
\hline
\rowcolor{LightCyan}
AP@.75 (APIoU=.75) & AP at IoU=.75 (strict metric) \\
\hline
\multicolumn{2}{l}{\textbf{Average Precision Across Scales}} \\
\hline
\rowcolor{LightCyan}
AP-S (APsmall) & AP for small objects: area $<$ $32^2$ \\
\hline
AP-M (APmedium) & AP for medium objects: $32^2$ $<$ area $<$ $96^2$ \\
\hline
\rowcolor{LightCyan}
AP-L (APlarge) & AP for large objects: area $>$ $96^2$ \\
\hline
\multicolumn{2}{l}{\textbf{Average Recall (AR)}} \\
\hline
\rowcolor{LightCyan}
AR1 (ARmax=1) & AR given 1 detection per image \\
\hline
AR10 (ARmax=10) & AR given 10 detections per image \\
\hline
\rowcolor{LightCyan}
AR100 (ARmax=100) & AR given 100 detections per image \\
\hline
\multicolumn{2}{l}{\textbf{Average Recall Across Scales}} \\
\hline
\rowcolor{LightCyan}
AR-S (ARsmall) & AR for small objects: area $<$ $32^2$ \\
\hline
AR-M (ARmedium) & AR for medium objects: $32^2$ $<$ area $<$ $96^2$ \\
\hline
\rowcolor{LightCyan}
AR-L (ARlarge) & AR for large objects: area $>$ $96^2$ \\
\hline
\end{tabular}
\label{tab_metrics}
\end{center}
\end{table}

In this study, we employ all of these metrics to compare performance between the competing methods.

\section{RESULTS AND DISCUSSIONS}
\label{sec_res_dis}

\subsection{Experimental Results on Test Sets of the Used Datasets}
\label{subsec_res}
In this section, we present results obtained using recent object detection approaches compared with existing methods in the human hand detection domain, e.g., the approaches implemented in \cite{yolo_github1}.

\subsubsection{Results on the EgoHands dataset}
The best results on this dataset are obtained from the YOLOv8n and YOLOv8x models, with AP measures of 0.768 and 0.766, respectively. The AR1, AR10, and AR100 metrics of these two approaches are also approximate. For YOLOv8n, the AR1 is 0.278 and for YOLOv8x, it is 0.277. The AR10 and AR100 scores of YOLOv8n are both 0.800, while those of YOLOv8x are 0.802. 

The existing approaches, i.e., YOLOv3 variants and YOLOv4-tiny employed in \cite{yolo_github1}, yield mediocre results. For instance, the AP metric for the normal variant of YOLOv3 is 0.659, with AR1 at 0.254, and both AR10 and AR100 at 0.698. The tiny variants of YOLOv3 and YOLOv4 (i.e., YOLOv3-tiny, YOLOv3-tiny-PRN, and YOLOv4-tiny) produce inferior results compared to the normal variant (YOLOv3). Notably, the YOLOv4-tiny variant implemented in \cite{yolo_github1} exhibits the poorest performance on this dataset, with an AP metric of 0.069, an AR1 of 0.073, and both AR10 and AR100 at 0.109.

It is important to note that the EgoHands dataset is part of the training data for both the YOLOv3 and YOLOv4 approaches employed in \cite{yolo_github1}. This inclusion is why the performance gap between the YOLOv3 approaches in \cite{yolo_github1} and the YOLOv8 models is relatively small. Nevertheless, the performance of the YOLOv4-tiny method is surprisingly poor, exhibiting a significant gap compared with the YOLOv8 models.

The DETA-swin-large model achieves the second-worst result on this dataset among the competing methods, with an AP metric of 0.248, an AR1 of 0.125, and AR10 and AR100 scores of 0.440 and 0.441, respectively. 

The other ViT-based method, i.e., the DETR-ResNet-50 model, produces reasonable results on this dataset, with its performance ranked third among competing methods, just behind the YOLOv8n and YOLOv8x models. Its AP measure is 0.702, while its AR1 is 0.258, and its AR10 and AR100 are both 0.760.

\begin{table*}[htbp]
\caption{Hand Detection Performance of Competing Approaches on the \textbf{EgoHands} Dataset}
\begin{center}
\begin{tabular}{|l|c|c|c|c|c|c|c|c|c|c|c|c|c|}
\hline
\textbf{Methods} & \textbf{AP} & \textbf{AP@.50} & \textbf{AP@.75} & \textbf{AP-S} & \textbf{AP-M} & \textbf{AP-L} & \textbf{AR1} & \textbf{AR10} & \textbf{AR100} & \textbf{AR-S} & \textbf{AR-M} & \textbf{AR-L} \\
\hline
\rowcolor{LightCyan}
YOLOv3-normal \cite{yolo_github1} & 0.659 & 0.940 & 0.814 & 0.042 & 0.238 & 0.706 & 0.254 & 0.698 & 0.698 & 0.040 & 0.270 & 0.747 \\
\hline
YOLOv3-tiny \cite{yolo_github1} & 0.531 & 0.888 & 0.602 & 0.035 & 0.090 & 0.579 & 0.225 & 0.577 & 0.577 & 0.033 & 0.113 & 0.629 \\
\hline
\rowcolor{LightCyan}
YOLOv3-tiny-PRN \cite{yolo_github1} & 0.503 & 0.877 & 0.548 & 0.000 & 0.060 & 0.552 & 0.217 & 0.555 & 0.555 & 0.000 & 0.087 & 0.608 \\
\hline
YOLOv4-tiny \cite{yolo_github1} & 0.069 & 0.180 & 0.033 & 0.000 & 0.020 & 0.085 & 0.073 & 0.109 & 0.109 & 0.000 & 0.040 & 0.117 \\
\hline
\rowcolor{LightCyan}
\textbf{YOLOv8n} & 0.768 & 0.969 & 0.899 & 0.134 & 0.471 & 0.798 & 0.278 & 0.800 & 0.800 & 0.227 & 0.519 & 0.834 \\
\hline
\textbf{YOLOv8x} & 0.766 & 0.969 & 0.903 & 0.164 & 0.493 & 0.796 & 0.277 & 0.802 & 0.802 & 0.193 & 0.545 & 0.833 \\
\hline
\textbf{DETR-ResNet-50} & 0.702 & 0.967 & 0.848 & 0.129 & 0.396 & 0.734 & 0.258 & 0.760 & 0.760 & 0.207 & 0.481 & 0.792 \\
\hline
\rowcolor{LightCyan}
\textbf{DETA-swin-large} & 0.248 & 0.628 & 0.125 & 0.000 & 0.018 & 0.287 & 0.125 & 0.440 & 0.441 & 0.000 & 0.090 & 0.481 \\
\hline
\end{tabular}
\label{tab_ego_hands}
\end{center}
\end{table*}

\subsubsection{Results on the 11k hands dataset}
This dataset presents the simplest scenario for hand detection, as each image features a clear single hand with high resolution. Consequently, the YOLOv8 models achieve nearly perfect results on this dataset, with the AP metric for YOLOv8n and YOLOv8x being 0.998 and 0.999, respectively. All the AR1, AR10, and AR100 metrics of these two models are perfect, each registering at 1.000. The gap between the approaches in \cite{yolo_github1} and the YOLOv8 models is huge on this dataset. This is understandable because the approaches in \cite{yolo_github1} were not trained on the 11k hands data. The normal YOLOv3 model achieves the maximum performance among the approaches in \cite{yolo_github1}, with an AP metric of 0.370, AR1 of 0.414, and both AR10 and AR100 at 0.415. Remarkably, the YOLOv4-tiny model fails to detect any hand objects.

The DETA-swin-large method achieves comparable results to the YOLOv8 models, with an AP metric of 0.996, and all of its AR1, AR10, and AR100 scores are 0.998. This performance is also comparable with that of the DETR-ResNet-50 method, which exhibits slightly lower AP measures than those of the DETA-swin-large method but slightly higher AR measures.

\begin{table*}[htbp]
\caption{Hand Detection Performance of Competing Approaches on the \textbf{11k Hands} Dataset}
\begin{center}
\begin{tabular}{|l|c|c|c|c|c|c|c|c|c|c|c|c|c|c|}
\hline
\textbf{Methods} & \textbf{AP} & \textbf{AP@.50} & \textbf{AP@.75} & \textbf{AP-S} & \textbf{AP-M} & \textbf{AP-L} & \textbf{AR1} & \textbf{AR10} & \textbf{AR100} & \textbf{AR-S} & \textbf{AR-M} & \textbf{AR-L}\\
\hline
\rowcolor{LightCyan}
YOLOv3-normal \cite{yolo_github1} & 0.370 & 0.674 & 0.381 & -1.000 & -1.000 & 0.371 & 0.414 & 0.415 & 0.415 & -1.000 & -1.000 & 0.415\\
\hline
YOLOv3-tiny \cite{yolo_github1} & 0.052 & 0.127 & 0.032 & -1.000 & -1.000 & 0.053 & 0.064 & 0.065 & 0.065 & -1.000 & -1.000 & 0.065\\
\hline
\rowcolor{LightCyan}
YOLOv3-tiny-PRN \cite{yolo_github1} & 0.007 & 0.029 & 0.000 & -1.000 & -1.000 & 0.007 & 0.010 & 0.010 & 0.010 & -1.000 & -1.000 & 0.010\\
\hline
YOLOv4-tiny \cite{yolo_github1} & 0.000 & 0.000 & 0.000 & -1.000 & -1.000 & 0.000 & 0.000 & 0.000 & 0.000 & -1.000 & -1.000 & 0.000\\
\hline
\rowcolor{LightCyan}
\textbf{YOLOv8n} & 0.998 & 1.000 & 1.000 & -1.000 & -1.000 & 0.998 & 1.000 & 1.000 & 1.000 & -1.000 & -1.000 & 1.000 \\
\hline
\textbf{YOLOv8x} & 0.999 & 1.000 & 1.000 & -1.000 & -1.000 & 0.999 & 1.000 & 1.000 & 1.000 & -1.000 & -1.000 & 1.000 \\
\hline
\rowcolor{LightCyan}
\textbf{DETR-ResNet-50} & 0.989 & 0.990 & 0.990 & -1.000 & -1.000 & 0.989 & 0.999 & 0.999 & 0.999 & -1.000 & -1.000 & 0.999 \\
\hline
\textbf{DETA-swin-large} & 0.996 & 1.000 & 1.000 & -1.000 & -1.000 & 0.996 & 0.998 & 0.998 & 0.998 & -1.000 & -1.000 & 0.998 \\
\hline
\end{tabular}
\label{tab_11k_hands}
\end{center}
\end{table*}

\subsubsection{Results on the Open Images dataset}
This dataset comprises hand images captured in various resolutions, conditions, and contexts, reflecting real-world scenarios. Among all methods evaluated, the YOLOv8x model emerges as the top performer, surpassing even the DETR-ResNet-50 and DETA-swin-large models. The YOLOv8x model achieves an AP metric of 0.417, with corresponding AR1, AR10, and AR100 scores of 0.252, 0.499, and 0.506, respectively. Following closely behind, both the DETR-ResNet-50 and YOLOv8n models rank as the second-best methods, each achieving an AP of 0.355. The AR1, AR10, and AR100 values of the DETR-ResNet-50 model are slightly higher than those of the YOLOv8n model. The former model achieves the AR1, AR10, and AR100 scores of 0.234, 0.466, and 0.478, respectively, while the latter model's scores are 0.233, 0.421, and 0.424, respectively.

The performance of the DETA-swin-large model ranks third in this dataset according to most evaluation metrics. When comparing the DETA-swin-large and YOLOv8n models, DETA-swin-large outperforms YOLOv8n in only two metrics: AP@.50 (0.678 vs 0.593) and AR-S (0.144 vs 0.142). However, the DETA-swin-large model is inferior to YOLOv8n in all other metrics.

The existing approaches based on YOLOv3 and YOLOv4 in \cite{yolo_github1} rank last among the competing methods, with the highest performance achieved by the normal YOLOv3. Its AP is 0.126, AR1 is 0.108, and both AR10 and AR100 are 0.159.

\begin{table*}[htbp]
\caption{Hand Detection Performance of Competing Approaches on the \textbf{Open Images} Dataset}
\begin{center}
\begin{tabular}{|l|c|c|c|c|c|c|c|c|c|c|c|c|c|}
\hline
\textbf{Methods} & \textbf{AP} & \textbf{AP@.50} & \textbf{AP@.75} & \textbf{AP-S} & \textbf{AP-M} & \textbf{AP-L} & \textbf{AR1} & \textbf{AR10} & \textbf{AR100} & \textbf{AR-S} & \textbf{AR-M} & \textbf{AR-L} \\

\hline
\rowcolor{LightCyan}
YOLOv3-normal \cite{yolo_github1} & 0.126 & 0.283 & 0.083 & 0.029 & 0.099 & 0.151 & 0.108 & 0.159 & 0.159 & 0.039 & 0.130 & 0.189 \\
\hline
YOLOv3-tiny \cite{yolo_github1} & 0.053 & 0.148 & 0.025 & 0.012 & 0.037 & 0.065 & 0.058 & 0.081 & 0.081 & 0.016 & 0.059 & 0.099 \\
\hline
\rowcolor{LightCyan}
YOLOv3-tiny-PRN \cite{yolo_github1} & 0.045 & 0.131 & 0.017 & 0.011 & 0.034 & 0.056 & 0.051 & 0.072 & 0.072 & 0.012 & 0.052 & 0.090 \\
\hline
YOLOv4-tiny \cite{yolo_github1} & 0.060 & 0.165 & 0.029 & 0.026 & 0.071 & 0.061 & 0.056 & 0.094 & 0.095 & 0.040 & 0.114 & 0.094 \\
\hline
\rowcolor{LightCyan}
\textbf{YOLOv8n} & 0.355 & 0.593 & 0.365 & 0.105 & 0.263 & 0.428 & 0.233 & 0.421 & 0.424 & 0.142 & 0.323 & 0.508 \\
\hline
\textbf{YOLOv8x} & 0.417 & 0.695 & 0.429 & 0.157 & 0.323 & 0.492 & 0.252 & 0.499 & 0.506 & 0.226 & 0.412 & 0.587 \\
\hline
\rowcolor{LightCyan}
\textbf{DETR-ResNet-50} & 0.355 & 0.702 & 0.315 & 0.118 & 0.268 & 0.432 & 0.234 & 0.466 & 0.478 & 0.207 & 0.369 & 0.564 \\
\hline
\textbf{DETA-swin-large} & 0.304 & 0.678 & 0.228 & 0.081 & 0.220 & 0.370 & 0.212 & 0.400 & 0.406 & 0.144 & 0.308 & 0.485 \\
\hline
\end{tabular}
\label{tab_open_images}
\end{center}
\end{table*}

\begin{table*}[htbp]
\caption{Hand Detection Performance of Competing Approaches on the \textbf{Combined} Dataset}
\begin{center}
\begin{tabular}{|l|c|c|c|c|c|c|c|c|c|c|c|c|c|}
\hline
\textbf{Methods} & \textbf{AP} & \textbf{AP@.50} & \textbf{AP@.75} & \textbf{AP-S} & \textbf{AP-M} & \textbf{AP-L} & \textbf{AR1} & \textbf{AR10} & \textbf{AR100} & \textbf{AR-S} & \textbf{AR-M} & \textbf{AR-L} \\
\hline
\rowcolor{LightCyan}
YOLOv3-normal \cite{yolo_github1} & 0.115 & 0.232 & 0.101 & 0.016 & 0.062 & 0.133 & 0.217 & 0.294 & 0.294 & 0.048 & 0.173 & 0.327 \\
\hline
YOLOv3-tiny \cite{yolo_github1} & 0.038 & 0.090 & 0.026 & 0.010 & 0.021 & 0.043 & 0.084 & 0.134 & 0.134 & 0.027 & 0.077 & 0.149 \\
\hline
\rowcolor{LightCyan}
YOLOv3-tiny-PRN \cite{yolo_github1} & 0.031 & 0.075 & 0.022 & 0.010 & 0.018 & 0.035 & 0.065 & 0.112 & 0.112 & 0.022 & 0.066 & 0.124 \\
\hline
YOLOv4-tiny \cite{yolo_github1} & 0.018 & 0.043 & 0.014 & 0.009 & 0.024 & 0.017 & 0.044 & 0.063 & 0.063 & 0.043 & 0.121 & 0.052 \\
\hline
\rowcolor{LightCyan}
\textbf{YOLOv8x (ego)} & 0.070 & 0.111 & 0.074 & 0.011 & 0.052 & 0.077 & 0.121 & 0.206 & 0.206 & 0.023 & 0.138 & 0.226 \\
\hline
\textbf{YOLOv8x (11k)} & 0.269 & 0.279 & 0.266 & 0.000 & 0.000 & 0.330 & 0.269 & 0.270 & 0.270 & 0.000 & 0.000 & 0.332 \\
\hline
\rowcolor{LightCyan}
\textbf{YOLOv8x (oi)} & 0.475 & 0.593 & 0.504 & 0.054 & 0.125 & 0.546 & 0.517 & 0.679 & 0.680 & 0.254 & 0.443 & 0.742 \\
\hline
\textbf{YOLOv8x (combined)} & \textbf{0.542} & \textbf{0.653} & \textbf{0.573} & \textbf{0.038} & \textbf{0.180} & \textbf{0.625} & \textbf{0.539} & \textbf{0.588} & \textbf{0.588} & \textbf{0.058} & \textbf{0.273} & \textbf{0.668} \\
\hline
\rowcolor{LightCyan}
\textbf{YOLOv8n (combined)} & 0.499 & 0.589 & 0.526 & 0.019 & 0.127 & 0.584 & 0.500 & 0.531 & 0.531 & 0.025 & 0.183 & 0.617 \\
\hline
\textbf{DETR-ResNet-50} & 0.515 & 0.667 & 0.520 & 0.035 & 0.137 & 0.600 & 0.524 & 0.581 & 0.581 & 0.114 & 0.275 & 0.658 \\
\hline
\rowcolor{LightCyan}
\textbf{DETA-swin-large} & 0.414 & 0.601 & 0.411 & 0.024 & 0.078 & 0.495 & 0.486 & 0.488 & 0.488 & 0.046 & 0.157 & 0.568 \\
\hline
\end{tabular}
\label{tab_all3_hands}
\end{center}
\end{table*}

\subsubsection{Results on the combined dataset}
The results on this combined dataset are comprehensive, as we evaluate the performance of models trained on individual datasets on the test set of this combined dataset. Among the YOLOv8x models, YOLOv8x trained on the EgoHands dataset, i.e., YOLOv8x (ego), exhibits the least optimal performance, with an AP of 0.070, AR1 of 0.121, AR10 and AR100 both of 0.206. In contrast, the YOLOv8x model trained on the combined dataset demonstrates the best performance, with an AP of 0.542, AR1 of 0.539, AR10 and AR100 both of 0.588. Additionally, the model trained on the Open Images dataset, i.e., YOLOv8x (oi), outperforms the model trained on the 11k dataset, i.e., YOLOv8x (11k), with an AP of 0.475 compared to 0.269. 

These comparison results demonstrate the impact of the training data and the similarity between the testing data and training data on the test performance of the object detection models. The YOLOv8x (ego) exhibits the worst performance, primarily due to the EgoHands dataset having the smallest number of training samples among the four datasets and contributing the least to the test set of the combined dataset.

The YOLOv8x (oi) model outperforms the YOLOv8x (11k) model due to the larger number of training samples in the Open Images dataset, which contributes more testing samples to the combined dataset compared to the 11k dataset. It is understandable that the YOLOv8x model trained on the combined dataset achieves the best performance. This is because it includes all available types of hand images in both the training and testing sets of the combined dataset.

A comparison between the nano (YOLOv8n) and extra large (YOLOv8x) variants shows that YOLOv8x is superior to YOLOv8n, with an AP metric of 0.542 compared to 0.499 for YOLOv8n. Similarly, the AR1 metric of YOLOv8x is 0.539, whereas it is 0.500 for YOLOv8n.

The DETR-ResNet-50 model ranks as the second-best method, just below the YOLOv8x model. It achieves an AP metric of 0.515, with AR1 at 0.524, and both AR10 and AR100 at 0.581. DETR-ResNet-50 outperforms YOLOv8x in three metrics: AP@.50, AR-S, and AR-M. Specifically, DETR's AP@.50 is 0.667 compared to YOLOv8x's 0.653, and its AR-S is 0.114, significantly higher than YOLOv8x's 0.058. Additionally, DETR's AR-M slightly surpasses that of YOLOv8x, with values of 0.275 versus 0.273, respectively.

Despite being the most computationally expensive model, DETA-swin-large performs worse than YOLOv8n and YOLOv8x on this dataset, consistent with its performance on other datasets. Its AP metric is 0.414, with AR1 at 0.486, and both AR10 and AR100 at 0.488.

The existing methods implemented in \cite{yolo_github1} based on YOLOv3 and YOLOv4 models exhibit notable inferiority compared to YOLOv8 models. The best method in \cite{yolo_github1} is based on the YOLOv3-normal model, with an AP of 0.115, AR1 of 0.217, and both AR10 and AR100 at 0.294. On the other hand, the YOLOv4-tiny model performs the worst among the competing methods, with an exceptionally low AP of 0.018, AR1 of 0.044, and both AR10 and AR100 at 0.063.

\subsection{Identifying Images/Frames with High Forensic Values}
In this section, we employ high-performing fine-tuned models to detect hand images within an extensive collection or video frames that may possess significant forensic significance. These images and frames could then be scrutinized further by forensic professionals in their analysis. We maintain a considerable margin for false positives, given that these images and frames will undergo thorough manual inspection by forensic experts subsequently. Based on the evaluation results obtained from the four datasets, we select the YOLOv8 model fine-tuned on the combined dataset for this purpose. Images or frames containing a significant portion of the hand are identified as having high forensic value. The size or area of the hand portion within an image can be calculated using the height and width of the bounding boxes detected by the object detection models. For a large folder of images, we predict on all images and filter out those with a hand area larger than a predetermined threshold value. The same process is applied to video frames. If an image contains more than one detected hand, the area of the largest hand is compared with the threshold. The threshold value can be adjusted; a higher threshold results in fewer images being selected, and vice versa.

\begin{figure}[htb]
\centerline{\includegraphics[width=1.0\linewidth]{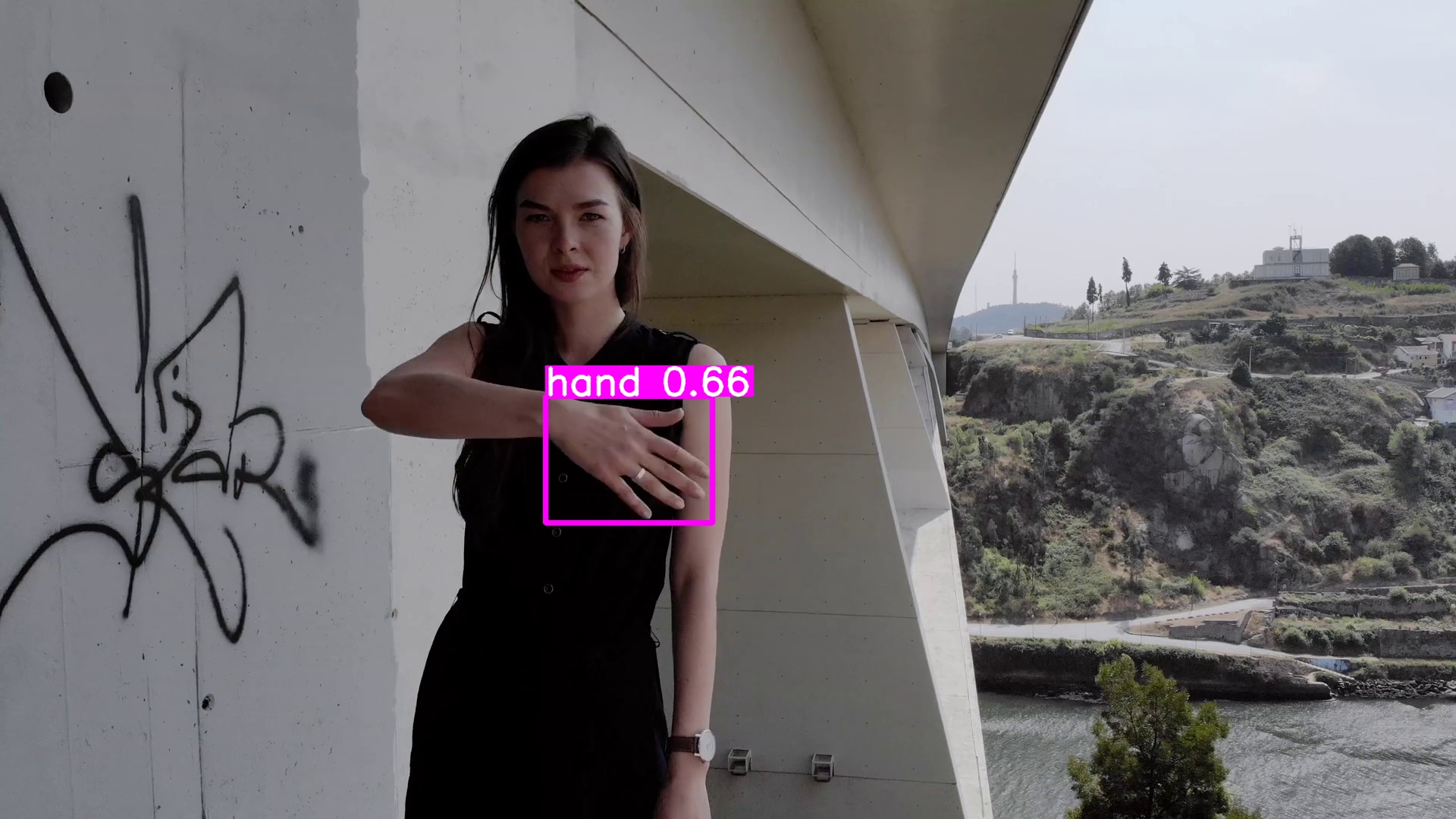}}
\caption{An example of a high forensic value frame obtained from the woman dancing video (https://youtu.be/CMq7392CIaY). This frame has a hand with its area larger than 30,000 square pixels, i.e., a threshold we selected for this video. We detect 8 frames in the video that are deemed having high forensic value, i.e., having a hand area larger than 30,000 square pixels. The rest 672 frames have a hand area less than 30,000 square pixels and thus are saved separately. The 8 frames above the threshold can be presented to forensic experts and this helps reduce their time in scanning the whole video.}
\label{fig:frame1}
\end{figure}

\begin{figure}[htb]
\centerline{\includegraphics[width=1.0\linewidth]{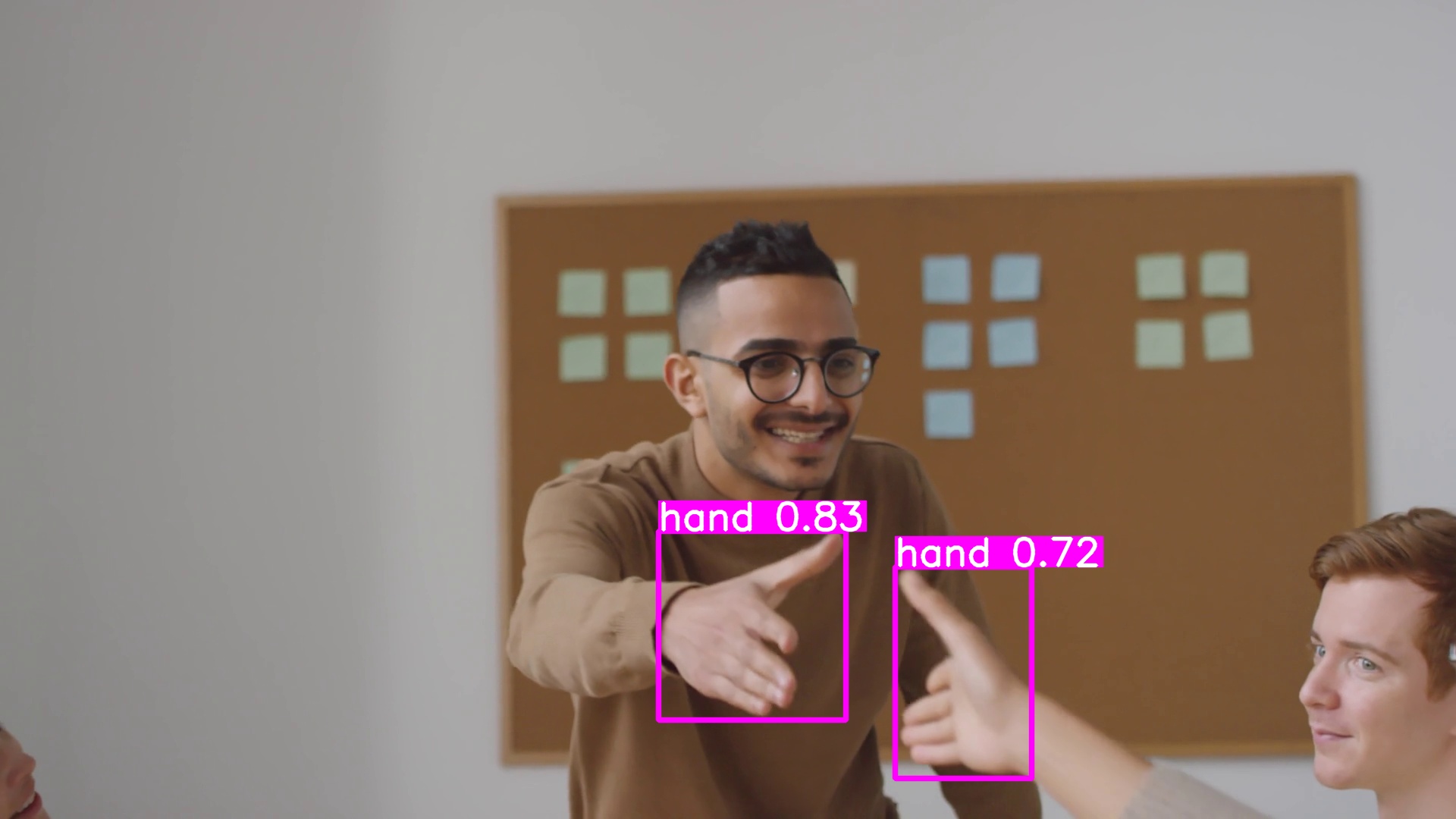}}
\caption{An example of a high forensic value frame obtained from the video of a man greeting a group in a business meeting (https://youtu.be/1G063siNpQ4). There are two hands detected in this frame. The larger hand has an area greater than 50,000 square pixels. With this threshold, 41 frames are detected as having high forensic value and the rest 456 frames are not.}
\label{fig:frame2}
\end{figure}

Table \ref{tab_videos} provides links to example videos processed by our fine-tuned YOLOv8 model. These videos were originally sourced from https://www.pexels.com/ and analyzed using the YOLOv8 hand detection pipeline to identify frames of high forensic value. Images with significant hand sizes or areas, and thus deemed of high forensic value, are saved separately, with several examples shown in Figs. \ref{fig:frame1}-\ref{fig:frame2}.

\begin{table}[htbp]
\caption{Example Videos Showing Results of Hand Object Detection}
\begin{center}
\begin{tabular}{ll}
\hline
\textbf{Video names} & \textbf{Links} \\
\hline
A woman dancing & https://youtu.be/CMq7392CIaY \\
\hline
Students raising hands & https://youtu.be/yqArrU1nnfI \\
\hline
People in a room & https://youtu.be/MAn1YiwjarM \\
\hline
People in a courtyard & https://youtu.be/7-234tg3JLc \\
\hline
People playing cards & https://youtu.be/ff\_fIRxz2-M \\
\hline
A man greeting a group & https://youtu.be/1G063siNpQ4 \\
\hline
Friends on truck & https://youtube.com/shorts/vj53Emxekfw \\
\hline
\end{tabular}
\label{tab_videos}
\end{center}
\end{table}

\section{CONCLUSION AND FURTHER WORK}
\label{sec_con_fut}
This paper presents object detection approaches based on YOLOv8 and detection transformers models for identifying hand images with high forensic value. We provided bounding box labels for the 11k hands dataset and created a combined dataset that includes hand images captured in various conditions and contexts. This enables us to effectively train hand object detection models and improve their performance compared to existing hand detection methods in the literature. We assessed different variants of YOLOv8 and detection transformers across four datasets, ultimately selecting the most effective approach for recognizing hand images with notable forensic relevance. The YOLOv8 models have shown superior performance compared to detection transformer methods, such as DETR-ResNet-50 and DETA-swin-large. The utilization of YOLOv8 models in forensic practice yields exceptional results, significantly reducing the time required by forensic experts. 

The hand region serves as a significant indicator of an image's forensic value. Nonetheless, images containing extensive hand areas can still encounter challenges such as occlusions, low-quality hand segments resulting from intense lighting, illumination effects, motion blur, or complex backgrounds. Further investigation is needed to address these factors and ultimately enhance efficiency, thereby reducing the workload for forensic experts.

Beyond the technical aspect, it is important to note the ethical and legal considerations of using hand images for identification \cite{deliversky2018ethical}. Privacy concerns, data protection, and informed consent are critical facets that need to be taken into account \cite{mordini2023biometric}. These aspects should be carefully considered in the future when implementing hand image-based identification systems in real-world applications.



\bibliographystyle{IEEEtran}
\bibliography{BibTexFile}

\begin{thebibliography}{10}
\providecommand{\url}[1]{#1}
\csname url@samestyle\endcsname
\providecommand{\newblock}{\relax}
\providecommand{\bibinfo}[2]{#2}
\providecommand{\BIBentrySTDinterwordspacing}{\spaceskip=0pt\relax}
\providecommand{\BIBentryALTinterwordstretchfactor}{4}
\providecommand{\BIBentryALTinterwordspacing}{\spaceskip=\fontdimen2\font plus
\BIBentryALTinterwordstretchfactor\fontdimen3\font minus \fontdimen4\font\relax}
\providecommand{\BIBforeignlanguage}[2]{{%
\expandafter\ifx\csname l@#1\endcsname\relax
\typeout{** WARNING: IEEEtran.bst: No hyphenation pattern has been}%
\typeout{** loaded for the language `#1'. Using the pattern for}%
\typeout{** the default language instead.}%
\else
\language=\csname l@#1\endcsname
\fi
#2}}
\providecommand{\BIBdecl}{\relax}
\BIBdecl

\bibitem{hackman2023forensic}
L.~Hackman and S.~Black, ``Forensic examination of the hand,'' \emph{Journal of the Royal Anthropological Institute}, vol.~29, pp. 116--131, 2023.

\bibitem{singla2020automated}
N.~Singla, M.~Kaur, and S.~Sofat, ``Automated latent fingerprint identification system: A review,'' \emph{Forensic Science International}, vol. 309, p. 110187, 2020.

\bibitem{badiye2023palmprints}
A.~Badiye, A.~Kamble, and N.~Kapoor, ``Palmprints: An introduction,'' in \emph{Textbook of Forensic Science}.\hskip 1em plus 0.5em minus 0.4em\relax Springer, 2023, pp. 279--293.

\bibitem{Zhang2021}
\BIBentryALTinterwordspacing
D.~Zhang and V.~Kanhangad, ``Hand geometry recognition,'' in \emph{Encyclopedia of Cryptography, Security and Privacy}, S.~Jajodia, P.~Samarati, and M.~Yung, Eds.\hskip 1em plus 0.5em minus 0.4em\relax Springer Berlin Heidelberg, 2021, pp. 1--4. [Online]. Available: \url{https://doi.org/10.1007/978-3-642-27739-9\_878-2}
\BIBentrySTDinterwordspacing

\bibitem{kuzu2022intra}
R.~S. Kuzu, E.~Maiorana, and P.~Campisi, ``On the intra-subject similarity of hand vein patterns in biometric recognition,'' \emph{Expert Systems with Applications}, vol. 192, p. 116305, 2022.

\bibitem{dahal2023interdisciplinary}
A.~Dahal, D.~McNevin, M.~Chikhani, and J.~Ward, ``An interdisciplinary forensic approach for human remains identification and missing persons investigations,'' \emph{Wiley Interdisciplinary Reviews: Forensic Science}, no. e1484, pp. 1--34, 2023.

\bibitem{soltani2019formal}
S.~Soltani and S.~A.~H. Seno, ``A formal model for event reconstruction in digital forensic investigation,'' \emph{Digital Investigation}, vol.~30, pp. 148--160, 2019.

\bibitem{zou2023object}
Z.~Zou, K.~Chen, Z.~Shi, Y.~Guo, and J.~Ye, ``Object detection in 20 years: A survey,'' \emph{Proceedings of the IEEE}, vol. 111, no.~3, pp. 257--276, 2023.

\bibitem{dhillon2020convolutional}
A.~Dhillon and G.~K. Verma, ``Convolutional neural network: a review of models, methodologies and applications to object detection,'' \emph{Progress in Artificial Intelligence}, vol.~9, no.~2, pp. 85--112, 2020.

\bibitem{sahin2023detection}
M.~E. Sahin, H.~Ulutas, E.~Yuce, and M.~F. Erkoc, ``Detection and classification of covid-19 by using faster r-cnn and mask r-cnn on ct images,'' \emph{Neural Computing and Applications}, vol.~35, no.~18, pp. 13\,597--13\,611, 2023.

\bibitem{diwan2023object}
T.~Diwan, G.~Anirudh, and J.~V. Tembhurne, ``Object detection using yolo: Challenges, architectural successors, datasets and applications,'' \emph{Multimedia Tools and Applications}, vol.~82, no.~6, pp. 9243--9275, 2023.

\bibitem{wang2023single}
X.~Wang, K.~Li, B.~Shi, L.~Li, H.~Lin, X.~Wang, and J.~Yang, ``Single shot multibox detector object detection based on attention mechanism and feature fusion,'' \emph{Journal of Electronic Imaging}, vol.~32, no.~2, pp. 023\,032--023\,032, 2023.

\bibitem{gehrig2023recurrent}
M.~Gehrig and D.~Scaramuzza, ``Recurrent vision transformers for object detection with event cameras,'' in \emph{Proceedings of the IEEE/CVF Conference on Computer Vision and Pattern Recognition}, 2023, pp. 13\,884--13\,893.

\bibitem{carion2020end}
N.~Carion, F.~Massa, G.~Synnaeve, N.~Usunier, A.~Kirillov, and S.~Zagoruyko, ``End-to-end object detection with transformers,'' in \emph{European Conference on Computer Vision}.\hskip 1em plus 0.5em minus 0.4em\relax Springer, 2020, pp. 213--229.

\bibitem{ouyang2022nms}
J.~Ouyang-Zhang, J.~H. Cho, X.~Zhou, and P.~Kr{\"a}henb{\"u}hl, ``{NMS} strikes back,'' \emph{arXiv preprint arXiv:2212.06137}, 2022.

\bibitem{afifi201911k}
M.~Afifi, ``11k hands: Gender recognition and biometric identification using a large dataset of hand images,'' \emph{Multimedia Tools and Applications}, vol.~78, pp. 20\,835--20\,854, 2019.

\bibitem{chao2021dexycb}
Y.-W. Chao, W.~Yang, Y.~Xiang, P.~Molchanov, A.~Handa, J.~Tremblay, Y.~S. Narang, K.~Van~Wyk, U.~Iqbal, S.~Birchfield \emph{et~al.}, ``{DexYCB}: A benchmark for capturing hand grasping of objects,'' in \emph{Proceedings of the IEEE/CVF Conference on Computer Vision and Pattern Recognition}, 2021, pp. 9044--9053.

\bibitem{spurr2021self}
A.~Spurr, A.~Dahiya, X.~Wang, X.~Zhang, and O.~Hilliges, ``Self-supervised {3D} hand pose estimation from monocular {RGB} via contrastive learning,'' in \emph{Proceedings of the IEEE/CVF International Conference on Computer Vision}, 2021, pp. 11\,230--11\,239.

\bibitem{lin2021end}
K.~Lin, L.~Wang, and Z.~Liu, ``End-to-end human pose and mesh reconstruction with transformers,'' in \emph{Proceedings of the IEEE/CVF Conference on Computer Vision and Pattern Recognition}, 2021, pp. 1954--1963.

\bibitem{shan2020understanding}
D.~Shan, J.~Geng, M.~Shu, and D.~F. Fouhey, ``Understanding human hands in contact at internet scale,'' in \emph{Proceedings of the IEEE/CVF Conference on Computer Vision and Pattern Recognition}, 2020, pp. 9869--9878.

\bibitem{joshi2020deep}
S.~V. Joshi and R.~D. Kanphade, ``Deep learning based person authentication using hand radiographs: A forensic approach,'' \emph{IEEE Access}, vol.~8, pp. 95\,424--95\,434, 2020.

\bibitem{narasimhaswamy2019contextual}
S.~Narasimhaswamy, Z.~Wei, Y.~Wang, J.~Zhang, and M.~Hoai, ``Contextual attention for hand detection in the wild,'' in \emph{Proceedings of the IEEE/CVF International Conference on Computer Vision}, 2019, pp. 9567--9576.

\bibitem{he2017mask}
K.~He, G.~Gkioxari, P.~Doll{\'a}r, and R.~Girshick, ``Mask {R-CNN},'' in \emph{Proceedings of the IEEE International Conference on Computer Vision}, 2017, pp. 2961--2969.

\bibitem{haji2023vision}
M.~N. Haji~Mohd, M.~S. Mohd~Asaari, O.~Lay~Ping, and B.~A. Rosdi, ``Vision-based hand detection and tracking using fusion of kernelized correlation filter and single-shot detection,'' \emph{Applied Sciences}, vol.~13, no.~13, p. 7433, 2023.

\bibitem{yolov8_ultralytics}
\BIBentryALTinterwordspacing
G.~Jocher, A.~Chaurasia, and J.~Qiu, ``Ultralytics {YOLOv8},'' 2023. [Online]. Available: \url{https://github.com/ultralytics/ultralytics}
\BIBentrySTDinterwordspacing

\bibitem{lin2015microsoft}
T.-Y. Lin, M.~Maire, S.~Belongie, L.~Bourdev, R.~Girshick, J.~Hays, P.~Perona, D.~Ramanan, C.~L. Zitnick, and P.~Dollár, ``Microsoft {COCO}: common objects in context,'' \url{https://arxiv.org/abs/1405.0312v3}, February 2015.

\bibitem{detr-resnet-50}
{Hugging Face}, ``{DETR} (end-to-end object detection) model with {ResNet-50} backbone,'' \url{https://huggingface.co/facebook/detr-resnet-50}, November 2023.

\bibitem{zhu2021deformable}
\BIBentryALTinterwordspacing
X.~Zhu, W.~Su, L.~Lu, B.~Li, X.~Wang, and J.~Dai, ``Deformable {DETR}: Deformable transformers for end-to-end object detection,'' in \emph{International Conference on Learning Representations}, 2021. [Online]. Available: \url{https://openreview.net/forum?id=gZ9hCDWe6ke}
\BIBentrySTDinterwordspacing

\bibitem{bambach2015lending}
S.~Bambach, S.~Lee, D.~J. Crandall, and C.~Yu, ``Lending a hand: Detecting hands and recognizing activities in complex egocentric interactions,'' in \emph{Proceedings of the IEEE International Conference on Computer Vision}, 2015, pp. 1949--1957.

\bibitem{kuznetsova2020open}
A.~Kuznetsova, H.~Rom, N.~Alldrin, J.~Uijlings, I.~Krasin, J.~Pont-Tuset, S.~Kamali, S.~Popov, M.~Malloci, A.~Kolesnikov \emph{et~al.}, ``The {Open Images Dataset V4}: Unified image classification, object detection, and visual relationship detection at scale,'' \emph{International Journal of Computer Vision}, vol. 128, no.~7, pp. 1956--1981, 2020.

\bibitem{OpenImages2}
\BIBentryALTinterwordspacing
I.~Krasin, T.~Duerig, N.~Alldrin, V.~Ferrari, S.~Abu-El-Haija, A.~Kuznetsova, H.~Rom, J.~Uijlings, S.~Popov, S.~Kamali, M.~Malloci, J.~Pont-Tuset, A.~Veit, S.~Belongie, V.~Gomes, A.~Gupta, C.~Sun, G.~Chechik, D.~Cai, Z.~Feng, D.~Narayanan, and K.~Murphy, ``{OpenImages}: A public dataset for large-scale multi-label and multi-class image classification.'' 2017. [Online]. Available: \url{https://storage.googleapis.com/openimages/web/index.html}
\BIBentrySTDinterwordspacing

\bibitem{open_od_leaderboard}
R.~Padilla, A.~Roberts, and the Hugging Face~Team, ``Open object detection leaderboard on {Hugging Face},'' \url{https://huggingface.co/spaces/rafaelpadilla/object\_detection\_leaderboard}, September 2023.

\bibitem{padilla2020survey}
R.~Padilla, S.~L. Netto, and E.~A. Da~Silva, ``A survey on performance metrics for object-detection algorithms,'' in \emph{International Conference on Systems, Signals and Image Processing (IWSSIP)}.\hskip 1em plus 0.5em minus 0.4em\relax IEEE, 2020, pp. 237--242.

\bibitem{yolo_github1}
F.~Bruggisser, ``A pre-trained {YOLO} based hand detection network,'' \url{https://github.com/cansik/yolo-hand-detection}, October 2022.

\bibitem{deliversky2018ethical}
J.~Deliversky and M.~Deliverska, ``Ethical and legal considerations in biometric data usage—{Bulgarian} perspective,'' \emph{Frontiers in Public Health}, vol.~6, no.~25, pp. 1--5, 2018.

\bibitem{mordini2023biometric}
E.~Mordini, ``Biometric privacy protection: What is this thing called privacy?'' \emph{IET Biometrics}, vol.~12, no.~4, pp. 183--193, 2023.

\end{thebibliography}

\end{document}